\pdfoutput=1

\documentclass[11pt]{article}

\usepackage{acl}

\usepackage{times}
\usepackage{latexsym}
\usepackage{multirow}
\usepackage{colortbl}
\usepackage{booktabs}
\usepackage{amssymb}
\usepackage{amsmath}
\usepackage{amsthm}
\usepackage{graphicx}
\usepackage[T1]{fontenc}

\usepackage[utf8]{inputenc}

\usepackage{microtype}
\title{Divide and Conquer: Text Semantic Matching \\ with Disentangled Keywords and Intents}

\author{{\normalsize Yicheng Zou$^{1,2}$,\ \ Hongwei Liu$^2$,\ \ Tao Gui$^1$\thanks{{ }{ }Corresponding authors.}\ \ ,\ \  Junzhe Wang$^2$,\ \ Qi Zhang$^2$$^*$,}\\
{\normalsize {\bf Meng Tang}$^3$,\ \ {\bf Haixiang Li}$^3$,\ \ {\bf Daniel Wang}$^3$}\\
  {$^1$\normalsize Institute of Modern Languages and Linguistics, Fudan University, Shanghai, China}\\
  {$^2$\normalsize School of Computer Science, Fudan University, Shanghai, China}\\
  {$^2$\normalsize Shanghai Collaborative Innovation Center of Intelligent Visual Computing, Shanghai, China}\\ 
  {$^3$\normalsize IPS, Tencent PCG, Beijing, China}\\
  \texttt{\normalsize \{yczou18,tgui,qz\}@fudan.edu.cn, daniellwang@tencent.com}}

\begin{document}
\maketitle
\begin{abstract}
Text semantic matching is a fundamental task that has been widely used in various scenarios, such as community question answering, information retrieval, and recommendation. Most state-of-the-art matching models, e.g., BERT, directly perform text comparison by processing each word uniformly. However, a query sentence generally comprises content that calls for different levels of matching granularity. Specifically, {\em keywords} represent factual information such as action, entity, and event that should be strictly matched, while {\em intents} convey abstract concepts and ideas that can be paraphrased into various expressions. In this work, we propose a simple yet effective training strategy for text semantic matching in a divide-and-conquer manner by disentangling keywords from intents. Our approach can be easily combined with pre-trained language models (PLM) without influencing their inference efficiency, achieving stable performance improvements against a wide range of PLMs on three benchmarks.
\end{abstract}

\section{Introduction}
Text semantic matching aims to predict a matching category or a matching score reflecting the semantic similarity given a pair of text sequences, which is a fundamental task employed in a wide range of applications \citep{huang2013learning,hu2014convolutional,palangi2016deep,cer2017semeval,ruckle2020multicqa,pang2021match}. Recently, pre-trained language models (PLM) show remarkable capability of representation learning and have accelerated the research of text semantic matching \citep{devlin2019bert,liu2019roberta,lan2019albert}. They typically exploit large-scale corpora and well-designed self-supervised learning objectives to better learn semantic representations, achieving state-of-the-art performances or even surpassing the level of non-expert humans on general semantic matching benchmarks \citep{wang2019glue,wang2019superglue}.
\begin{figure}
\centering
  \includegraphics[width=3.0in]{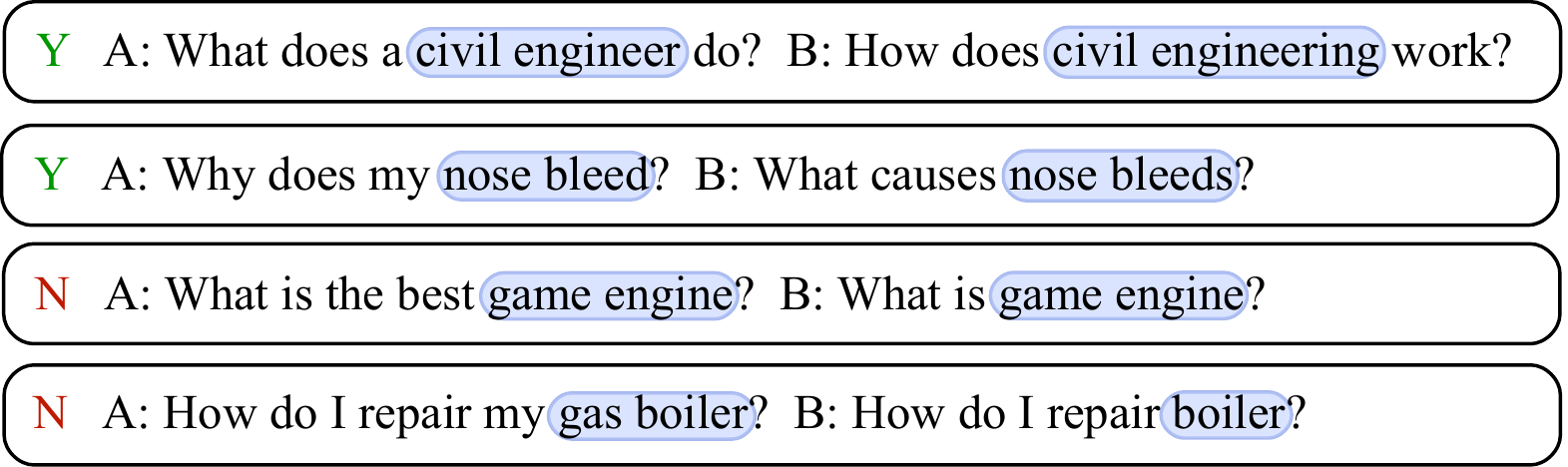}
  \caption{Examples of sentence pairs sampled from the QQP dataset. The {\bf keywords} are highlighted, while the other words constitute abstract {\bf intents}. {\bf Y} and {\bf N} represent whether the pair is matched or not. The original matching problem can be decomposed into two sub-problems: keyword matching and intent matching. A semantically equivalent pair generally means the keyword and intent are matched simultaneously.}  \label{fig:intro}
\end{figure}

Most existing PLMs aim to establish a foundation for various downstream tasks \citep{bommasani2021opportunities} and focus on finding a generic way to encode text sequences. When applied to the task of text semantic matching, it is a common practice to add a simple classification objective for fine-tuning and directly perform text comparison by processing each word uniformly. Nevertheless, each sentence can be typically decomposed into content with different levels of matching granularity \citep{su2021keep}. Exemplar sentence pairs can be found in Figure \ref{fig:intro}. The primary content refers to {\em keywords} that reflect the factual information about entities or actions, which should be strictly matched. The other content constitute abstract {\em intents}, which can be generally paraphrased into various expressions to convey the same concepts or ideas.

Considering the situation where sentence content has different levels of matching granularity, we propose {\bf DC-Match}, a simple but effective training regime for text semantic matching in a divide-and-conquer manner. Specifically, we break down the matching problem into two sub-problems: {\em keyword matching} and {\em intent matching}. Given a pair of input text sequences, the model learns to disentangle keywords from intents by utilizing the method of distant supervision. In addition to the standard sequence matching that has a global receptive field, we further match keywords and intents separately to learn the way of content matching under different levels of granularity. Finally, we design a special training objective that combines the solutions to the sub-problems, which minimizes the KL-divergence between the global matching distribution (original problem) and the joint keyword-intent matching distribution (sub-problems). At inference time, we expect that the global matching model automatically distinguishes keywords from intents, then makes final predictions conditioned on the disentangled content in different matching levels.

We adopted DC-Match to a wide range of PLMs. Comprehensive experiments were conducted on two English text matching benchmarks MRPC \citep{dolan2005automatically} and QQP \citep{iyer2017first}, and a Chinese benchmark Medical-SM. Our approach can be easily combined with PLMs plus few additional parameters, but still achieves stable performance improvements against most baseline PLMs. Notably, all the auxiliary procedures and parameters are only involved in the training stage. The inference efficiency of our approach is exactly the same as that of PLM baselines, without additional parameters and computations. Our codes and datasets are publicly available\footnote{\url{https://github.com/RowitZou/DC-Match}}. 

Our contributions are three-fold: 1) We introduce a novel training regime for text matching, which disentangles keywords from intents based on different levels of matching granularity  in a divide-and-conquer manner. 2) The proposed approach is simple yet effective, which can be easily combined with PLMs plus few auxiliary training parameters while not changing their original inference efficiency. 3) Experimental results on three benchmarks across two languages demonstrate the effectiveness of our approach in different aspects.

\section{Related Work}
Text semantic matching plays an important role in many applications, such as Information Retrieval (IR) and Natural Language Inference (NLI). Traditional technologies exploit neural networks with different inductive biases, e.g., CNN \citep{tan2016improved}, RNN \citep{tai2015improved,cheng2016long}, GNN \cite{wu2020joint}, and attention mechanism \citep{parikh2016decomposable,chen2017enhanced}. To enhance the matching performance, dozens of works use richer syntactic or hand-crafted features \citep{chen2017enhanced,tay2018compare,gong2018natural,kim2019semantic}, add complex alignment computations \citep{wang2017bilateral,tan2018multiway,gong2018natural,yang2019simple}, and perform multi-pass matching procedures \citep{tay2018co,kim2019semantic}, which shows the effectiveness of representation-oriented approaches and model designing strategies based on information interaction. 

Recently, large-scale pre-trained language models (PLM) have boosted the performance of text semantic matching by making full use of massive text resources. Most of them are composed of multiple transformer layers \citep{vaswani2017attention} with multi-head attentions and are pre-trained with well-designed self-supervised learning objectives. Representative models like BERT \citep{devlin2019bert}, RoBERTa \citep{liu2019roberta}, and ALBERT \citep{lan2019albert} aim to establish a powerful encoder that has a comprehensive understanding of input texts. For the task of text semantic matching, PLMs can be fine-tuned under a paradigm of sequence classification with only an additional classification layer, achieving state-of-the-art performances on general semantic matching benchmarks \citep{wang2019glue,wang2019superglue}.
\begin{figure*}
\centering
  \includegraphics[width=5.8in]{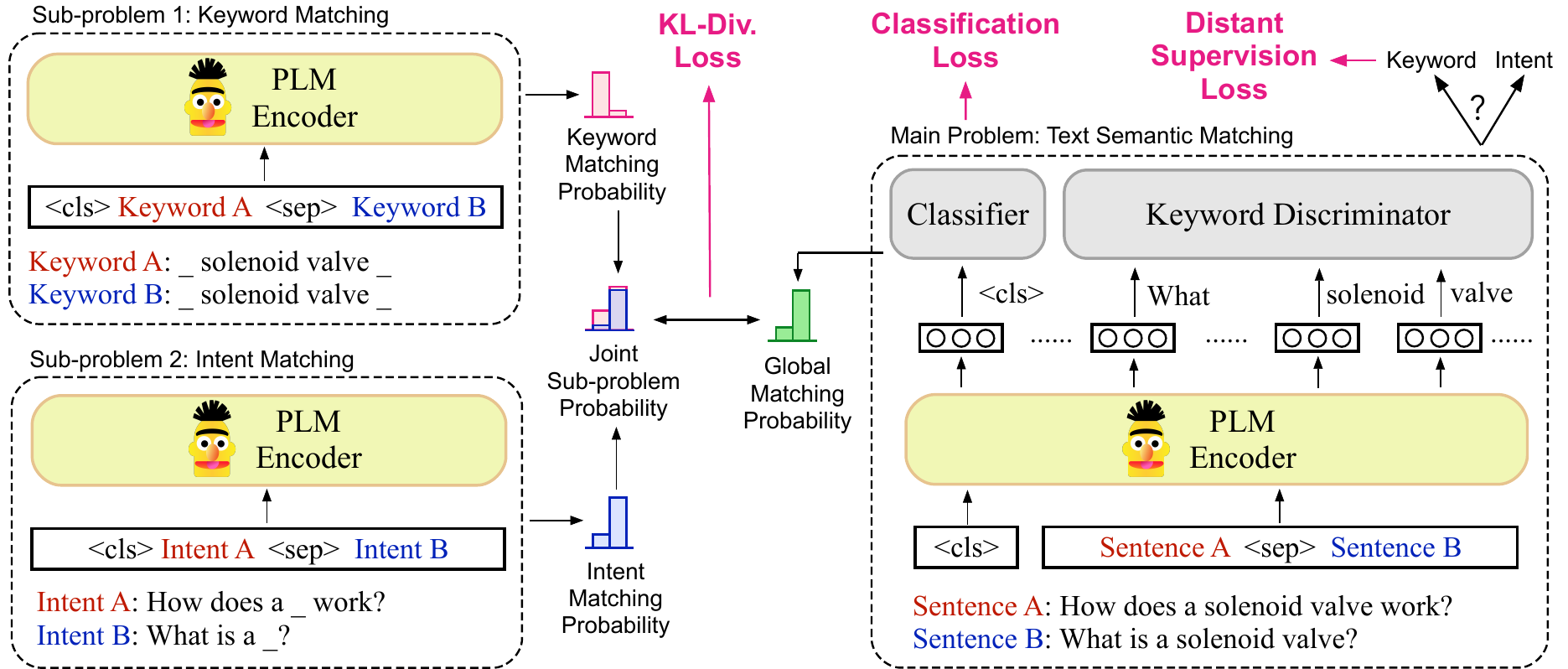}
  \caption{Overview of DC-Match. The training regime has three objectives: (1) a standard matching classification loss; (2) a distant supervision loss for keyword and intent discrimination; (3) a KL-divergence loss that makes the global matching probability (main problem) consistent with the probability of combined solutions to keyword matching and intent matching (sub-problems).} \label{fig:model}
\end{figure*}
PLMs can be regarded as foundation models \citep{bommasani2021opportunities} and they mainly focus on finding a generic way to encode text sequences. Instead of processing each word uniformly, in this work, we devise a novel training regime that processes sentence pairs by disentangling keywords from intents, which can be easily combined with PLMs to stack additional improvements to text semantic matching.

\vspace{-2.15pt}
\section{Methodology}
The proposed training regime DC-Match consists of three training objectives: a classification loss for the global matching model; a distantly supervised classification loss that learns to distinguish keywords from intents; a special training objective following the idea of divide and conquer, which uses the KL-divergence to ensure that the global matching distribution (original problem) is similar to the distribution of combined solutions to disentangled keywords and intents (sub-problems). The overall framework is illustrated in Figure \ref{fig:model}.

\subsection{Text Semantic Matching using PLMs}
\label{sec:plm}
First, we formally define the task of text semantic matching and describe a generic way for this task by using PLMs. Given two text sequences $S^a=\{w^a_1, w^a_2,...,w^a_{l_a}\}$ and $S^b=\{w^b_1, w^b_2,...,w^b_{l_b}\}$, the goal of text semantic matching is to learn a classifier $y=\xi(S^a, S^b)$ to predict whether $S^a$ and $S^b$ is semantically equivalent. Here, $w^a_i$ and $w^b_j$ represent the $i$-th and $j$-th word in the sequences, respectively, and $l_a$, $l_b$ denote the sequence length. $y$ can be either a binary classification target indicating whether or not the two sequences are matched, or a multi-class classification target that reflects different matching degrees.

Recently, pre-trained language models (PLM) have achieved remarkable success in text understanding and representation learning \citep{devlin2019bert,liu2019roberta,lan2019albert}.
They are pre-trained on large-scale text corpora with heuristic self-supervised learning objectives, and can be served as a powerful sequence classifier by fine-tuning on the downstream classification task. For text semantic matching, it is a common practice that we directly concatenate $S^a$ and $S^b$ as a consecutive sequence $S^{a,b}=[S^a;w^{sep};S^b]$ by a separator token $w^{sep}$ and feed it into the PLM encoder:
\begin{align}
    &[\mathbf{h}^{cls};\mathbf{H}^{a,b}] = \mathrm{PLM}([w^{cls};S^{a,b}]),
    \label{eq1}\\
    &P(y|S^a, S^b) = \mathrm{Softmax}(\mathbf{h}^{cls}\cdot \mathbf{W}^{\top}).
    \label{eq2}
\end{align}
Here, $w^{cls}$ is a special token in front of each sequence, and the final hidden state corresponding to this token $\mathbf{h}^{cls}$ is used as the aggregate sequence representation. During fine-tuning, only a single classification layer is introduced to make the final prediction, where $\mathbf{W}\in \mathbb{R}^{K\times H}$ represents trainable weights and $K$ is the number of labels. Finally, we compute a standard classification loss for fine-tuning as follows:
\begin{equation}
    \mathcal{L}_{sm} = -\mathrm{log}P(y|S^a, S^b).
    \label{eq3}
\end{equation}

\subsection{Disentangling Keyword from Intent with Distant Supervision}
Most existing PLMs aim to find a generic way to encode text sequences and establish a foundation for language understanding. For different classification tasks, e.g., sentiment analysis, text semantic matching, and natural language inference, the PLM typically exploits the same fine-tuning paradigm, and processes text sequences in a straightforward and uniform way. In this work, inspired by previous works of decomposable paraphrase generation \cite{li2019decomposable,su2021keep}, we introduce a task-specific assumption to the text semantic matching, and postulate that each sentence could be decomposed into keywords and intents. Intuitively, keywords represent factual information such as actions and entities that should be strictly matched, while intents convey abstract concepts or ideas that can be expressed in different ways. By disentangling keywords from intents, the matching procedure can be divided into two easier sub-problems that call for different levels of matching granularity.

However, automatic disentanglement of keywords and intents is not easy due to the lack of manually annotated data. To address this problem, following recent research on distant supervision \citep{liang2020bond,meng2021distantly}, we use a rule-based method to automatically generate keyword labels by extracting entity mentions in the raw text based on the entities in external knowledge bases (see details in Section \ref{sec:process}). All extracted entities are labeled as keywords and the remainder of the sentence words are labeled as intents. After obtaining the weakly labeled information, we add an auxiliary training objective that forces the model to learn disentangled keyword and intent representations. Formally, given the output states $\mathbf{H}^{a,b}$ from PLM in Eq.\ref{eq1}, we split the states into two groups $\mathbf{H}^{a,b}_{k}\in \mathbb{R}^{N_k \times H}$ and $\mathbf{H}^{a,b}_{i}\in \mathbb{R}^{N_i \times H}$ that correspond to the tokens of keywords and intents, respectively, where $N_k,N_i$ denote the token number. Then, the keyword-intent classification loss is defined as follows:
\begin{equation}
    \mathcal{L}_{ds} = -[\mathrm{log}\sigma(\hat{\mathbf{h}}^{a,b}_{k}\mathbf{W}_{ds}^{\top}) + \mathrm{log}\sigma(-\hat{\mathbf{h}}^{a,b}_{i}\mathbf{W}_{ds}^{\top})],
    \label{eq4}
\end{equation}
where $\mathbf{W}_{ds}\in \mathbb{R}^{1\times H}$ is trainable parameters, and $\hat{\mathbf{h}}^{a,b}_{k},\hat{\mathbf{h}}^{a,b}_{i}$ are transformed by $\mathbf{H}^{a,b}_{k},\mathbf{H}^{a,b}_{i}$ using average pooling. The objective in Eq.\ref{eq4} aims to push the encoder to learn representations of keywords and intents such that they are far apart from each other, modeling disentangled sentence content in different matching levels.

\subsection{Divide-and-Conquer Matching Strategy}
The auxiliary training objective in Eq.\ref{eq4}, nevertheless, cannot be directly associated with the original text matching problem. To facilitate the true contributions of keywords and intents to the final prediction, we introduce a novel matching strategy following the idea of divide and conquer. Specifically, we divide the original matching problem into two easier sub-problems: keyword matching and intent matching, and assume that they are independent to each other. The solutions to the sub-problems are then combined to give a solution to the original problem. Recall that the goal of text semantic matching is to learn $y=\xi(S^a, S^b)$ where $y$ can be either a binary classification target or a multi-class classification target. We assume that each sub-problem follows the same type of target, and the probability distribution of combined solutions $Q(y)$ can be derived from the joint probability distribution of the two sub-problems $P(y_k, y_i)$ as:
\begin{align}
    Q(y=c_n) &= \ \  P(y_k=c_n, y_i=c_n)\nonumber\\
    + &\sum\nolimits_{c_m>c_n} P(y_k=c_n, y_i=c_m) \nonumber\\
    + &\sum\nolimits_{c_m>c_n} P(y_k=c_m, y_i=c_n).
    \label{eq5}
\end{align}
Here, $c_n,c_m$ denote the target classes which reflect the matching degrees, and $c_m>c_n$ means $c_m$ has a higher matching score than $c_n$. For example, in a three-class scenario, $y\in\{2,1,0\}$ means exact match, partial match, and mismatch, respectively, and $Q(y=0)$ is the probability that at least one of the sub-problems is inferred as mismatched.

To model the sub-problems, we reuse the matching model in Eq.\ref{eq1} and Eq.\ref{eq2} to separately compare keywords and intents and get conditional probabilities $P(y_k|S^a_k,S^b_k)$ and $P(y_i|S^a_i,S^b_i)$. $S_k$ and $S_i$ represent text sequences where tokens of intents or keywords are masked, respectively. Then, under the assumption of independent sub-problems, the conditional joint distribution of $y_k$ and $y_i$ is:
\begin{equation}
    P(y_k,y_i|S^a,S^b) = P(y_k|S^a_k,S^b_k)P(y_i|S^a_i,S^b_i).
\end{equation}

Finally, we can combine the solutions to the sub-problems and compute the conditional distribution $Q(y|S^a,S^b)$ using Eq.\ref{eq5}. To ensure that the global matching distribution (original problem) is similar to the distribution of combined solutions to sub-problems, we use the bidirectional KL-divergence loss to minimize the distance between $P(y|S^a,S^b)$ and $Q(y|S^a,S^b)$ as follows:
\begin{align}
    \mathcal{L}_{dc}&=1/2\cdot(D_{KL}[P(y|S^a,S^b)||Q(y|S^a,S^b)]\nonumber\\
    &+D_{KL}[Q(y|S^a,S^b)||P(y|S^a,S^b)]).
    \label{eq7}
\end{align}
By this means, we expect that the global matching model learns to make final predictions with better interpretability, which are conditioned on the disentangled keywords and intents that require different levels of matching granularity.
\subsection{Training and Inference}
At the training stage, we combine the three loss functions $\mathcal{L}_{sm},\mathcal{L}_{ds},\mathcal{L}_{dc}$ to jointly train the model:
\begin{equation}
    \mathcal{L} = \mathcal{L}_{sm} + \mathcal{L}_{ds} + \mathcal{L}_{dc}.
\end{equation}
At the inference time, we directly infer the matching category for a sentence pair based on the conditional probability of the original problem, namely $y^*=\mathrm{argmax}_yP(y|S^a,S^b)$. It means our inference procedure is exactly the same as that of PLM baselines without additional computations. Here, we do not infer matching results from the probability of combined solutions $Q(y|S^a,S^b)$, since annotation information of keywords and intents is generally not available at the inference time, and $Q(y|S^a,S^b)$ cannot be directly computed. Although we use external corpora to automatically obtain distant labels, it might induce incomplete and noisy signals \citep{meng2021distantly}, introducing biases to $Q(y|S^a,S^b)$ approximation. Hence, we only use distant labels at the training stage as auxiliary information augmentation to the global matching model. Nevertheless, we observe that after model training, $P(y|S^a,S^b)$ is highly consistent with $Q(y|S^a,S^b)$ (see details in Section \ref{sec:analysis}). As a result, a high-quality set of keyword labels might bring additional performance enhancement by better approximating $Q(y|S^a,S^b)$.

\section{Experimental Settings}
\subsection{Datasets}
\begin{table}[t!]
\footnotesize
\begin{center}
\setlength{\tabcolsep}{2mm}{
\begin{tabular}{cccccc}
\toprule[1pt]
 \multirow{2}{*}{{\bf Split}}&\bf \# of& \bf Avg. & \multicolumn{3}{c}{\bf \# of pairs in categories}\\
 & \bf pairs & \bf length & EM(2) & PM(1) & MM(0)\\
\midrule
Train & 38,406 & 12.25 & 7,754  & 18,617& 12,035\\
Dev. & 4,801 & 12.25 & 975  & 2,329 & 1,497 \\
Test & 4,801 & 12.19 & 938 & 2,315 & 1,548 \\
\bottomrule[1pt]
\end{tabular}}
\end{center}
\caption{\label{tb:data}Statistics of the Medical-SM dataset. Each query pair can be categorized into exact match (EM), partial match (PM), or mismatch (MM).}
\end{table}
We evaluate our approach and all baselines on three benchmarks for text semantic matching: two English datasets MRPC \citep{dolan2005automatically} and QQP \citep{iyer2017first}, and one Chinese dataset Medical-SM. Both MRPC and QQP are corpora of sentence pairs automatically extracted from online websites, with annotated binary classification labels indicating whether the sentences in the pair are semantically equivalent. We use the official dataset collections on Glue \citep{wang2019glue} released by the community\footnote{\url{https://huggingface.co/datasets/glue}}, where MRPC contains 5,801 sentence pairs and QQP consists of 404,276 annotated sentence pairs\footnote{Since the labels for the official QQP test set are not released, we report evaluation results on the validation set.}.

Furthermore, we evaluate our approach on a Chinese text matching dataset Medical-SM, which consists of user-generated query pairs collected from a Chinese search engine. The dataset  contains 48,008 query pairs in the domain of medical consulting. Each query pair can be categorized into three classes: exact match, partial match, or mismatch. The annotation is completed by five independent experts and we keep the labeling choices that most annotators accept. Statistics of our constructed dataset are shown in Table \ref{tb:data}. To facilitate the research, we will release the dataset publicly.

\begin{table}[t!]
\footnotesize
\begin{center}
\setlength{\tabcolsep}{2mm}{
\begin{tabular}{lccc}
\toprule[1pt]
 &{\bf QQP} &{\bf MRPC} & {\bf Medical}\\
\midrule
\# keywords in each pair &2.38 &6.53& 2.51\\
\# tokens in each keyword & 1.98&1.68& 4.51\\
BLEU (match) &.1451 &.3088&.2754\\
BLEU (mismatch) & .0961&.2155&.1284\\
\bottomrule[1pt]
\end{tabular}}
\end{center}
\caption{\label{tb:keyword}Statistics of distantly labeled keywords on training sets. BLEU (match/mismatch) denotes the keyword BLEU score in matched/mismatched pairs, respectively.}
\end{table}

\subsection{Automatic Keyword Labeling}
\label{sec:process}
In this work, we generate distant supervision labels for identification of keywords and intents using a heuristic approach. Inspired by previous works for distantly supervised NER \citep{liang2020bond,meng2021distantly}, we first extract potential keywords with part-of-speech tags of nouns, verbs, and adjectives obtained from NLTK \citep{bird2006nltk}. We then match these potential keywords by using external knowledge bases: wikipedia entity graph \citep{bhatia2018know} for English corpora, and Sogou knowledge graph \cite{wang2019knowledge} for Chinese Medical-SM. Finally, we use the binary IO format to label whether a token belongs to keywords or intents \citep{peng2019distantly}. Table \ref{tb:keyword} shows the statistics of distantly labeled keywords on the training sets of three benchmarks. We use BLEU score \citep{papineni2002bleu} to measure the relevance of keywords between two compared sentences for both matched pairs and mismatched pairs. We observe that matched sentence pairs generally contain keywords with higher relevance. As a result, generic models might wrongly output high matching scores just conditioned on matched keywords regardless of their context, because models tend to learn statistical biases in the data \citep{manjunatha2019explicit,lin2021using}. 


\subsection{Implementation Details}
For a fair comparison, we fine-tune each PLM of the original version and its DC-Match variant with the same set of hyper-parameters. The fine-tuning process of the QQP and MRPC datasets follows \citet{wang2021textflint}. Specifically, we apply AdamW \cite{loshchilov2018decoupled} ($\beta_1$=0.9, $\beta_2$=0.999) with a weight decay rate of 0.01 and set the learning rate to 2e-5. The batch size is set to 64 for QQP and 16 for MRPC. All experiments are conducted on a single RTX 3090 GPU. For QQP, we fine-tune the model for 50,000 steps and model checkpoints are evaluated every 2,000 steps. For MRPC, we fine-tune the model for 20 epochs and evaluate the model after each epoch. Checkpoints with top-3 performance on the development set are evaluated on the test set to report average results. For Medical-SM, we use the same fine-tuning strategy as for QQP, and use the chinese version of PLM checkpoints released by \citet{cui2021pretrain}\footnote{Since the large version of Chinese BERT is not available, we use Chinese MacBERT \citep{cui2020revisiting} instead of BERT. }.

\section{Results and Analysis}

\begin{table}[t!]
\footnotesize
\begin{center}
\setlength{\tabcolsep}{1.5mm}{
\begin{tabular}{|l|c|c|}
\hline
{\bf Model} & {\bf\ \ QQP\ \ } & {\bf MRPC}\\
\hline
CENN \citep{zhang2017context} & 80.7& 76.4\\
L.D.C \citep{wang2016sentence}& 85.6&78.4\\
BiMPM \citep{wang2017bilateral}& 88.2& -\\
DIIN \citep{gong2018natural}& 89.1& -\\
DRCN \citep{kim2019semantic}& 90.2  & 82.5 \\ 
DRr-Net \citep{zhang2019drr}& 89.8  & 82.9 \\
R$^2$-Net \citep{zhang2021making}& 91.6  & 84.3 \\
\hline
BERT \citep{devlin2019bert}& 90.9 & 82.7 \\
\quad -{\em large version}& 91.0 & 85.9 \\ 
RoBERTa \citep{liu2019roberta}& 91.4 &87.2 \\ 
\quad-{\em large version}& 92.0 &87.6 \\
ALBERT \citep{lan2019albert}& 90.4 & 86.0 \\
\quad -{\em large version}& 90.9 &86.5\\
DeBERTa \citep{he2020deberta}&91.7 & 88.4 \\
\quad-{\em large version}& 92.1& 88.6  \\
FunnelTF \citep{dai2020funnel}& 91.9 & 87.1\\
\hline
DC-Match (RoBERTa-base) & 91.7& 88.1\\
DC-Match (RoBERTa-large) & \bf 92.2 &\bf 88.9 \\
\hline
\end{tabular}}
\end{center}
\caption{\label{tb:main}Experimental results ({\bf Accuracy}) on the {\bf QQP} and {\bf MRPC} text semantic matching datasets. }
\end{table}

\subsection{Main Results}
Table \ref{tb:main} shows the main results of comparison models on the QQP and MRPC dataset. Following previous works \cite{zhang2021making,wang2021textflint}, we evaluate matching performance using {\bf Accuracy} and some results are from their reported scores. In Table \ref{tb:main}, all baselines are categorized into two groups. The first group includes traditional methods that exploit neural networks with different inductive biases, and the second group includes PLMs that benefit from large-scale external pre-training data. Unsurprisingly, PLMs show a superior performance against traditional neural matching models, especially on the small-scale dataset MRPC. When equipped with the DC-Match training strategy, PLMs can achieve further performance enhancement. In Table \ref{tb:main}, we report the results of DC-Match that uses RoBERTa as the backbone PLM, which outperforms all baselines on both datasets. However, the improvement on a single PLM does not necessarily mean the effect of DC-Match has generalizability. Hence, to probe the effectiveness of our proposed training regime, we apply DC-Match to all the PLMs in the second group and report the results of performance change in Table \ref{tb:qqp_mrpc}. Notably, the listed PLMs generally have different architectures and parameter scales, and we fine-tune each PLM of the original version and its DC-Match variant using the same set of configurations without additional tuning of hyper-parameters. We are surprised to find that the matching accuracy of all PLMs increases stably on both datasets. It indicates that the divide-and-conquer strategy by breaking down the matching problem into easier sub-problems can effectively give a better solution to the original problem. Besides, from Table \ref{tb:qqp_mrpc} we observe that DC-Match brings more significant performance boost to the small dataset MRPC, which probes that the information of keywords and intents is an important feature for text semantic matching, especially when the training data is too limited to find useful latent patterns. 

\begin{table}[t!]
\footnotesize
\begin{center}
\setlength{\tabcolsep}{1.5mm}{
\begin{tabular}{lcc}
\toprule[1pt]
\multirow{2}{*}{{\bf Model}} & {\bf QQP} & {\bf MRPC}\\
& Ori. $\to$ DC (change) & Ori. $\to$ DC (change)\\
\midrule
BERT & 90.91 $\to$ 91.16 ({\bf 0.25})& 82.66 $\to$ 83.82 ({\bf 1.16})\\
\quad -{\em large}& 90.98 $\to$ 91.45 ({\bf 0.47})& 85.85 $\to$ 86.08 (0.23)\\ 
RoBERTa & 91.41 $\to$ 91.69 ({\bf 0.28})&87.24 $\to$ 88.05 ({\bf 0.81})\\ 
\quad-{\em large}& 92.03 $\to$ 92.20 ({\bf 0.17})&87.59 $\to$ 88.92 ({\bf 1.33})\\
ALBERT & 90.37 $\to$ 90.62 ({\bf 0.25})& 86.02 $\to$ 86.26 (0.24)\\
\quad -{\em large}& 90.91 $\to$ 90.94 (0.03) & 86.49 $\to$ 87.01 ({\bf 0.52})\\
DeBERTa &91.68 $\to$ 91.78 (0.10)& 88.40 $\to$ 88.81 ({\bf 0.41})\\
\quad-{\em large}& 92.13 $\to$ 92.22 (0.09)& 88.57 $\to$ 89.21 ({\bf 0.64})\\
FunnelTF & 91.92 $\to$ 92.09 ({\bf 0.17})& 87.07 $\to$ 87.53 ({\bf 0.46})\\
\bottomrule[1pt]
\end{tabular}}
\end{center}
\caption{\label{tb:qqp_mrpc}Experimental results of  {\bf Accuracy} on the {\bf QQP} and {\bf MRPC} datasets. We compare the results of original PLMs with those using our DC-Match training strategy (Ori.$\to$DC), and calculate the improvement of accuracy. Numbers in {\bf bold} indicate whether the change is significant (using a Wilcoxon signed-rank test; $p<0.05$). }
\end{table}
Furthermore, we evaluate DC-Match on the Chinese Medical-SM. Different from QQP and MRPC, Medical-SM is a three-class classification dataset. In addition to accuracy, we further employ Macro-F1 to assess the quality of problems with multiple classes. From Table \ref{tb:medical} we observe that DC-Match still boosts the matching performance of PLMs, indicating that our strategy works fine in a multi-class classification scenario and in different languages.

\begin{table}[t!]
\footnotesize
\begin{center}
\setlength{\tabcolsep}{1.5mm}{
\begin{tabular}{lcc}
\toprule[1pt]
\multirow{2}{*}{{\bf Model}} & {\bf Accuracy} & {\bf Macro-F1}\\
& Ori. $\to$ DC (change) & Ori. $\to$ DC (change)\\
\midrule
BERT & 73.55 $\to$ 73.83 ({\bf 0.28})& 72.91 $\to$ 73.15 ({\bf 0.24})\\
\quad-{\em large}& 74.55 $\to$ 74.69 (0.14)&74.01 $\to$ 74.13 (0.12)\\
RoBERTa & 73.19 $\to$ 73.73 ({\bf 0.54})&72.43 $\to$ 72.96 ({\bf 0.53})\\ 
\quad-{\em large}& 73.51 $\to$ 74.22 ({\bf 0.71})&72.83 $\to$ 73.67 ({\bf 0.84})\\
\bottomrule[1pt]
\end{tabular}}
\end{center}
\caption{\label{tb:medical}{\bf Accuracy} and {\bf Macro-F1} on the {\bf Medical-SM} corpus. Numbers in {\bf bold} indicate the result change is significant (Wilcoxon signed-rank test; $p<0.05$). }
\end{table}
\begin{figure}
\centering
  \includegraphics[width=3.0in]{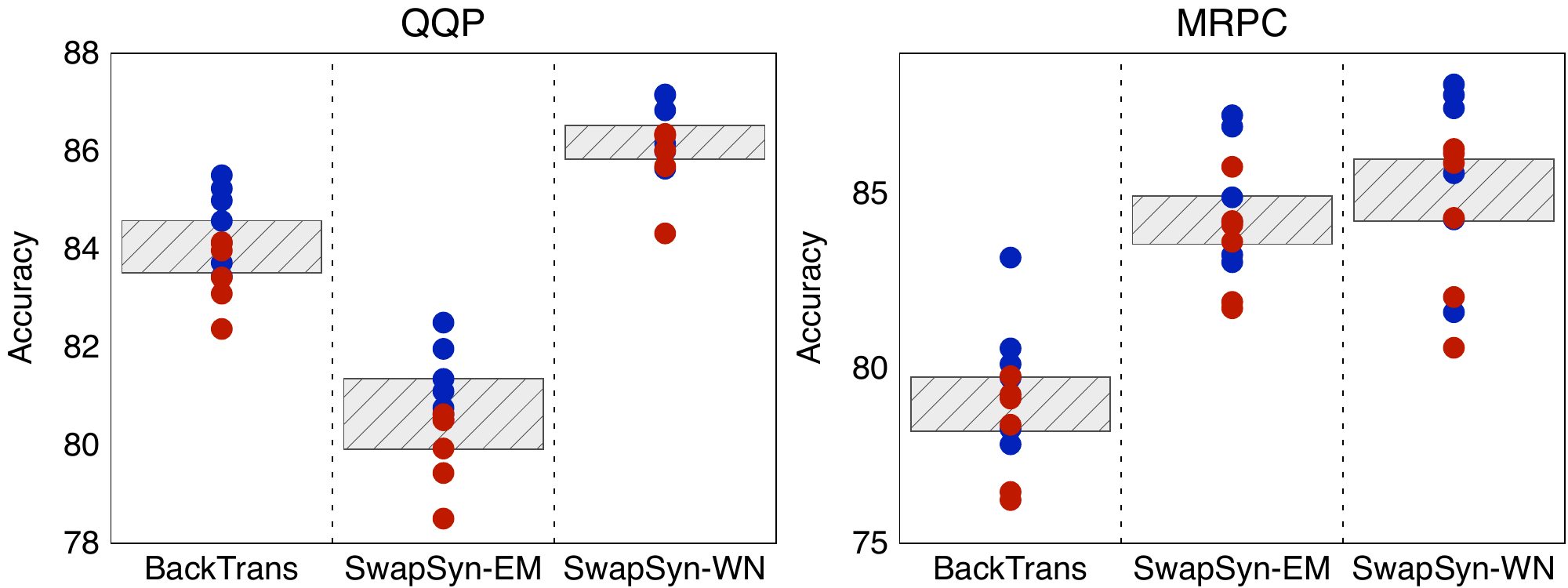}
  \caption{Robustness evaluation on the QQP and MRPC datasets. The x-axis denotes different text transformations that aim to test whether models are vulnerable to attacks. The y-axis denotes model accuracy on the transformed test set. {\color{red}Red} dots represent the original PLMs while {\color{blue} Blue} dots represent those using the DC-Match strategy. Bar plots denote the gap of mean accuracy between two groups of models.} \label{fig:robust}
\end{figure}
\subsection{Ablation Experiments}
We also perform ablation studies to validate the effectiveness of each part in DC-Match. Table \ref{tb:ablation} demonstrates the results of different settings for the proposed training strategy equipped with RoBERTa. After only adding the distantly supervised loss for keyword and intent identification (+$\mathcal{L}_{ds}$), we find that the results are not significantly different from the original PLMs. It reflects that this auxiliary training objective cannot be directly associated with the original text matching problem, so $\mathcal{L}_{ds}$ itself might not be helpful for the final target. However, if we remove $\mathcal{L}_{ds}$ from DC-Match and only keep the divide-and-conquer training objective (+$\mathcal{L}_{dc}$), we observe a performance degradation compared with the full version of DC-Match. It indicates that the distant supervision target helps the model learn to disentangle keywords from intents and obtain distinguished content representations that call for different levels of matching granularity, which might contribute to the solutions to sub-problems. Besides, the incorporation of the divide-and-conquer objective (both +$\mathcal{L}_{dc}$ and +$\mathcal{L}_{ds},\mathcal{L}_{dc}$) improves the performance of PLMs to varying degrees, which manifests the effectiveness of the matching strategy in a decomposed manner.  

\begin{table}[t!]
\small
\begin{center}
\setlength{\tabcolsep}{2mm}{
\begin{tabular}{lccc}
\toprule[1pt]
\bf Models & \bf QQP & \bf MRPC & \bf Medical-SM\\
\midrule
RoBERTa-base & 91.41 & 87.24 &73.19\\
\ \ + $\mathcal{L}_{ds}$  & 91.48 & 87.36& 73.30\\
\ \ + $\mathcal{L}_{dc}$ & 91.61& 87.88& 73.65\\
\ \ + $\mathcal{L}_{ds},\mathcal{L}_{dc}$ (ours) &91.69& 88.05 & 73.73 \\
\midrule
RoBERTa-large & 92.03 & 87.59 & 73.51 \\
\ \ + $\mathcal{L}_{ds}$ & 91.96& 87.86& 73.85\\
\ \ + $\mathcal{L}_{dc}$ & 92.15& 88.82& 74.13\\
\ \ + $\mathcal{L}_{ds},\mathcal{L}_{dc}$ (ours)& 92.20 & 88.92& 74.22\\
\bottomrule[1pt]
\end{tabular}}
\end{center}
\caption{\label{tb:ablation}Ablation study of DC-Match on three text semantic matching datasets. We report results of {\bf Accuracy} and use RoBERTa as the backbone model.}
\end{table}
\subsection{Robustness Evaluation}
The divide-and-conquer strategy disentangles keywords from intents, which provides additional interpretability for final matching judgements. Following \citet{wang2021textflint}, we conduct robustness evaluation to probe whether DC-Match is robust to text transformations by breaking down the matching problem into easier sub-problems. Specifically, we use a public toolkit\footnote{\url{https://www.textflint.io}} and test the following transformations: (1) {\bf BackTrans} transforms each sentence into a semantically equal sentence using back translation. (2) {\bf SwapSyn-WN} replaces words with synonyms provided by WordNet \cite{miller1995wordnet}. (3) {\bf SwapSyn-EM} replaces common words with synonyms using Glove Embeddings \citep{pennington2014glove}. We test 6 PLMs (BERT, ALBERT, RoBERTa with base and large version) in their original and DC-Match enhanced version, and report the results in Figure \ref{fig:robust}\footnote{All transformations are conducted on the subset of the original evaluation set where both the original PLMs and the DC-Match enhanced variants give accurate predictions.}. We observe that both original PLMs and their DC-Match variants suffer performance degradation. However, the DC-Match enhanced PLMs can keep a more stable performance compared to original ones, which manifests that DC-Match can improve the robustness of PLMs to a certain extent for the text semantic matching task.

\begin{table*}[t]
\small
\begin{center}
\setlength{\tabcolsep}{2mm}{
\begin{tabular}{lccccc}
\toprule[1pt]
\bf Sentence Pair& \bf Label &
\bf PLM &\bf DC&\bf Kw.&\bf In.\\
\midrule
A: What is the difference between an {\color{red}animal cell} and a {\color{red}plant cell}? & \multirow{2}{*}{0} & \multirow{2}{*}{1} & \multirow{2}{*}{0} & \multirow{2}{*}{0} & \multirow{2}{*}{1}\\
B: What is the difference between {\color{red}plant cell vacuoles} and {\color{red}animal cell vacuoles}? \\
\midrule
A: {\color{red}Benchmark Treasury 10-year notes} gained 17/32, yielding 4.015 percent. & \multirow{2}{*}{0} & \multirow{2}{*}{1} & \multirow{2}{*}{0} & \multirow{2}{*}{1} & \multirow{2}{*}{0}\\
B: The {\color{red}benchmark 10-year note} was recently down 17/32, to yield 4.067 percent.
 \\
\midrule
A: Is there any {\color{red}culture} difference between {\color{red}US} and {\color{red}UK}? & 
\multirow{2}{*}{1} & \multirow{2}{*}{0} & \multirow{2}{*}{1} & \multirow{2}{*}{1} & \multirow{2}{*}{1}\\
B: What is the biggest difference in {\color{red}British culture} and {\color{red}American culture}?\\
\midrule
A: But the {\color{red}cancer society} said its study had been misused. & 
\multirow{2}{*}{0} & \multirow{2}{*}{1} & \multirow{2}{*}{0} & \multirow{2}{*}{0} & \multirow{2}{*}{0}\\
B: The {\color{red} American Cancer Society} said the study was flawed in several ways.\\
\bottomrule[1pt]
\end{tabular}}
\end{center}
\caption{\label{tb:case} Test cases on the QQP and MRPC datasets. We use BERT-base as the backbone model. Words in {\color{red}Red} represent distantly labeled keywords. {\bf PLM}, {\bf DC}, {\bf Kw.}, and {\bf In.} represent predictions from the original PLMs, the DC-Match enhanced PLMs, and the DC-Match sub-problems (keyword matching and intent matching), respectively. }
\end{table*}
\subsection{Analysis of Divide-and-Conquer Strategy}
\label{sec:analysis}
\begin{figure}
\centering
  \includegraphics[width=3.0in]{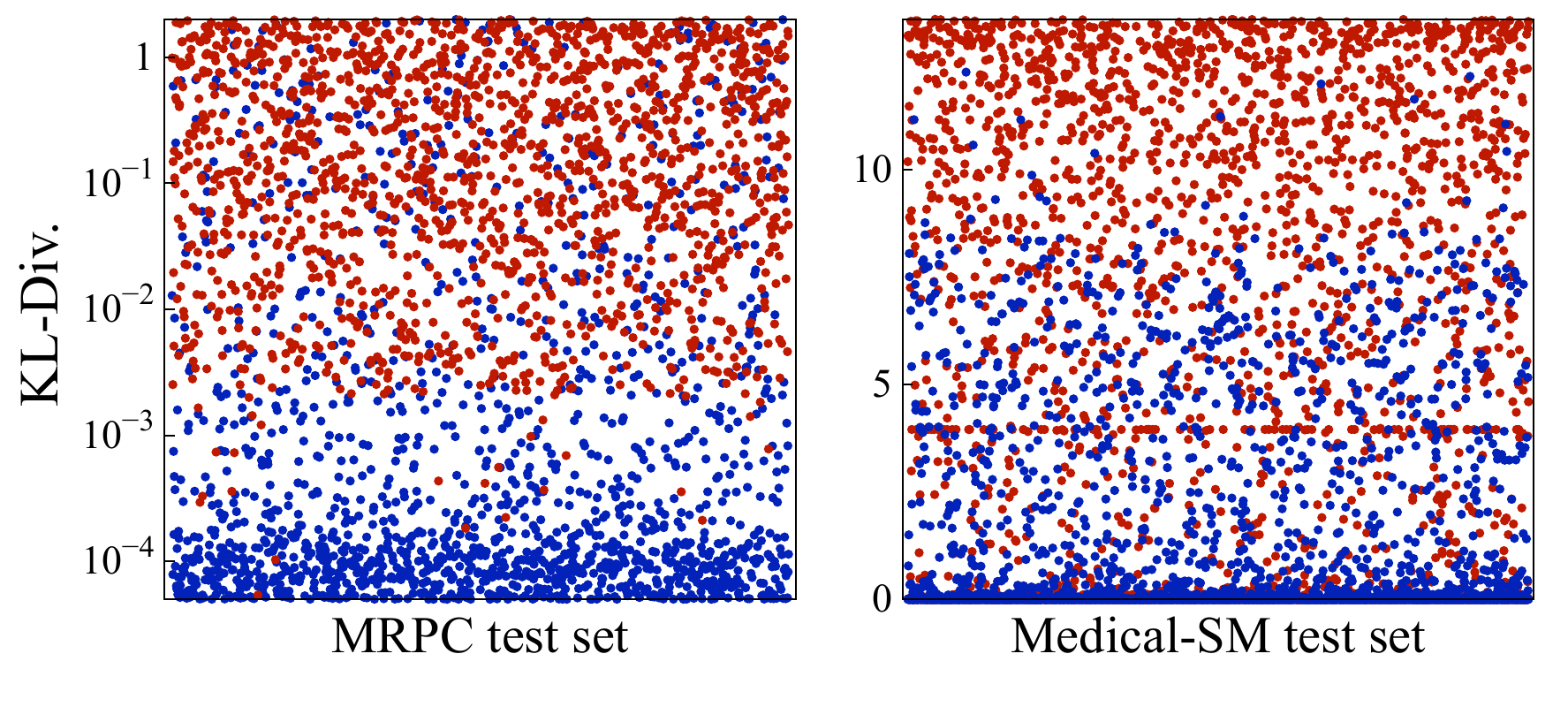}
  \caption{KL-divergence between $P(y)$ and $Q(y)$. Each point denotes the KL-divergence score of a test sample (1725 samples for MRPC and 4801 samples for Medical-SM). {\color{red}Red} dots are scores from the original PLMs, while {\color{blue}Blue} dots are those from DC-Match. BERT-base is used as the backbone model. We observe that DC-Match significantly narrows the gap between $P(y)$ and $Q(y)$ compared to the original PLMs. } \label{fig:dc}
\end{figure}

Recall that the model cannot access the labeled keywords at test time, so the probability of combined solutions to the sub-problems $Q(y)$ cannot be directly computed. Hence, the KL-divergence loss in Eq.\ref{eq7} is designed to minimize the distance between $Q(y)$ and the global matching probability $P(y)$, aiming to simulate the divide-and-conquer process at inference time. To probe that $P(y)$ can truly approximate $Q(y)$, we further label the keywords in test sets as described in Section \ref{sec:process}, so that we can calculate $Q(y)$ directly\footnote{Here, we exploit the keyword labels in test sets only for analysis, and they do not influence model predictions. }. We compute the KL-divergence score between $P(y)$ and $Q(y)$ for each test example and illustrate the results in Figure \ref{fig:dc}. Red dots denote scores from the original PLMs, while blue dots are scores from DC-Match. We can observe that $P(y)$ and $Q(y)$ show much higher consistency (lower KL-Div. scores) when using the DC-Match strategy compared to the original PLMs, which again manifests the effectiveness of our devised divide-and-conquer training objective that narrows the gap between $P(y)$ and $Q(y)$.

\subsection{Case Study}
To intuitively understand how the DC-Match strategy works, we show several test cases of the QQP and MRPC datasets with predicted labels from different systems in Table \ref{tb:case}. In order to analyze how the DC-Match enhanced PLMs make accurate predictions, we also show the solutions to the two sub-problems, namely $P(y_k|S^a_k,S^b_k)$ and $P(y_i|S^a_i,S^b_i)$, by directly introducing distant keyword labels as in Section \ref{sec:analysis}. From the cases we observe that the final predictions of DC-Match are highly consistent with those of sub-problems. The model tends to output a low matching score as long as at least one of the sub-problems is inferred as mismatched. We also find that the original PLMs tend to make wrong predictions when two mismatched sentences share long common sub-sequences. For example, in the first case, the main difference between two sentences is the concept of 'cell' and 'cell vacuoles', but the remainder of the sequences is almost the same, which might confuse the model. By contrast, DC-Match is capable of identifying keywords from text sequences, and can make accurate judgements by dividing the matching problem into easier sub-problems.

\section{Conclusion}
In this work, we devise a divide-and-conquer training strategy DC-Match for text semantic matching. It breaks down the matching problem into two sub-problems: keyword matching and intent matching. The model learns to disentangle keywords from intents that require different levels of matching granularity. The proposed DC-Match is simple and effective, which can be easily combined with PLMs plus few additional parameters. We conduct experiments on three text matching datasets across different languages. Experimental results show that our approach can not only achieve stable performance improvement, but also shows robustness to semantically invariant text transformations. 

\section*{Acknowledgments}
The authors wish to thank the anonymous reviewers for their helpful comments. This work was partially funded by  National Natural Science Foundation of China (No. 61976056, 62076069).
\bibliography{acl}

\begin{thebibliography}{51}
\expandafter\ifx\csname natexlab\endcsname\relax\def\natexlab#1{#1}\fi

\bibitem[{Bhatia and Vishwakarma(2018)}]{bhatia2018know}
Sumit Bhatia and Harit Vishwakarma. 2018.
\newblock Know thy neighbors, and more! studying the role of context in entity
  recommendation.
\newblock In \emph{Proceedings of the 29th on Hypertext and Social Media},
  pages 87--95.

\bibitem[{Bird(2006)}]{bird2006nltk}
Steven Bird. 2006.
\newblock Nltk: the natural language toolkit.
\newblock In \emph{Proceedings of the COLING/ACL 2006 Interactive Presentation
  Sessions}, pages 69--72.

\bibitem[{Bommasani et~al.(2021)Bommasani, Hudson, Adeli, Altman, Arora, von
  Arx, Bernstein, Bohg, Bosselut, Brunskill
  et~al.}]{bommasani2021opportunities}
Rishi Bommasani, Drew~A Hudson, Ehsan Adeli, Russ Altman, Simran Arora, Sydney
  von Arx, Michael~S Bernstein, Jeannette Bohg, Antoine Bosselut, Emma
  Brunskill, et~al. 2021.
\newblock On the opportunities and risks of foundation models.
\newblock \emph{arXiv preprint arXiv:2108.07258}.

\bibitem[{Cer et~al.(2017)Cer, Diab, Agirre, Lopez-Gazpio, and
  Specia}]{cer2017semeval}
Daniel Cer, Mona Diab, Eneko Agirre, I{\~n}igo Lopez-Gazpio, and Lucia Specia.
  2017.
\newblock Semeval-2017 task 1: Semantic textual similarity multilingual and
  crosslingual focused evaluation.
\newblock In \emph{Proceedings of the 11th International Workshop on Semantic
  Evaluation (SemEval-2017)}, pages 1--14.

\bibitem[{Chen et~al.(2017)Chen, Zhu, Ling, Wei, Jiang, and
  Inkpen}]{chen2017enhanced}
Qian Chen, Xiaodan Zhu, Zhen-Hua Ling, Si~Wei, Hui Jiang, and Diana Inkpen.
  2017.
\newblock Enhanced lstm for natural language inference.
\newblock In \emph{Proceedings of the 55th Annual Meeting of the Association
  for Computational Linguistics (Volume 1: Long Papers)}, pages 1657--1668.

\bibitem[{Cheng et~al.(2016)Cheng, Dong, and Lapata}]{cheng2016long}
Jianpeng Cheng, Li~Dong, and Mirella Lapata. 2016.
\newblock Long short-term memory-networks for machine reading.
\newblock In \emph{Proceedings of the 2016 Conference on Empirical Methods in
  Natural Language Processing}, pages 551--561.

\bibitem[{Cui et~al.(2020)Cui, Che, Liu, Qin, Wang, and Hu}]{cui2020revisiting}
Yiming Cui, Wanxiang Che, Ting Liu, Bing Qin, Shijin Wang, and Guoping Hu.
  2020.
\newblock Revisiting pre-trained models for chinese natural language
  processing.
\newblock In \emph{Proceedings of the 2020 Conference on Empirical Methods in
  Natural Language Processing: Findings}, pages 657--668.

\bibitem[{Cui et~al.(2021)Cui, Che, Liu, Qin, and Yang}]{cui2021pretrain}
Yiming Cui, Wanxiang Che, Ting Liu, Bing Qin, and Ziqing Yang. 2021.
\newblock Pre-training with whole word masking for chinese bert.
\newblock \emph{IEEE Transactions on Audio, Speech and Language Processing}.

\bibitem[{Dai et~al.(2020)Dai, Lai, Yang, and Le}]{dai2020funnel}
Zihang Dai, Guokun Lai, Yiming Yang, and Quoc Le. 2020.
\newblock Funnel-transformer: Filtering out sequential redundancy for efficient
  language processing.
\newblock In \emph{NeurIPS}.

\bibitem[{Devlin et~al.(2019)Devlin, Chang, Lee, and
  Toutanova}]{devlin2019bert}
Jacob Devlin, Ming-Wei Chang, Kenton Lee, and Kristina Toutanova. 2019.
\newblock Bert: Pre-training of deep bidirectional transformers for language
  understanding.
\newblock In \emph{Proceedings of the 2019 Conference of the North American
  Chapter of the Association for Computational Linguistics: Human Language
  Technologies, Volume 1 (Long and Short Papers)}, pages 4171--4186.

\bibitem[{Dolan and Brockett(2005)}]{dolan2005automatically}
William~B Dolan and Chris Brockett. 2005.
\newblock Automatically constructing a corpus of sentential paraphrases.
\newblock In \emph{Proceedings of the Third International Workshop on
  Paraphrasing (IWP2005)}.

\bibitem[{Gong et~al.(2018)Gong, Luo, and Zhang}]{gong2018natural}
Yichen Gong, Heng Luo, and Jian Zhang. 2018.
\newblock Natural language inference over interaction space.
\newblock In \emph{International Conference on Learning Representations}.

\bibitem[{He et~al.(2020)He, Liu, Gao, and Chen}]{he2020deberta}
Pengcheng He, Xiaodong Liu, Jianfeng Gao, and Weizhu Chen. 2020.
\newblock Deberta: Decoding-enhanced bert with disentangled attention.
\newblock In \emph{International Conference on Learning Representations}.

\bibitem[{Hu et~al.(2014)Hu, Lu, Li, and Chen}]{hu2014convolutional}
Baotian Hu, Zhengdong Lu, Hang Li, and Qingcai Chen. 2014.
\newblock Convolutional neural network architectures for matching natural
  language sentences.
\newblock In \emph{Advances in neural information processing systems}, pages
  2042--2050.

\bibitem[{Huang et~al.(2013)Huang, He, Gao, Deng, Acero, and
  Heck}]{huang2013learning}
Po-Sen Huang, Xiaodong He, Jianfeng Gao, Li~Deng, Alex Acero, and Larry Heck.
  2013.
\newblock Learning deep structured semantic models for web search using
  clickthrough data.
\newblock In \emph{Proceedings of the 22nd ACM international conference on
  Information \& Knowledge Management}, pages 2333--2338.

\bibitem[{Iyer et~al.(2017)Iyer, Dandekar, Csernai et~al.}]{iyer2017first}
Shankar Iyer, Nikhil Dandekar, Korn{\'e}l Csernai, et~al. 2017.
\newblock First quora dataset release: Question pairs.
\newblock \emph{data. quora. com}.

\bibitem[{Kim et~al.(2019)Kim, Kang, and Kwak}]{kim2019semantic}
Seonhoon Kim, Inho Kang, and Nojun Kwak. 2019.
\newblock Semantic sentence matching with densely-connected recurrent and
  co-attentive information.
\newblock In \emph{Proceedings of the AAAI conference on artificial
  intelligence}, volume~33, pages 6586--6593.

\bibitem[{Lan et~al.(2019)Lan, Chen, Goodman, Gimpel, Sharma, and
  Soricut}]{lan2019albert}
Zhenzhong Lan, Mingda Chen, Sebastian Goodman, Kevin Gimpel, Piyush Sharma, and
  Radu Soricut. 2019.
\newblock Albert: A lite bert for self-supervised learning of language
  representations.
\newblock In \emph{International Conference on Learning Representations}.

\bibitem[{Li et~al.(2019)Li, Jiang, Shang, and Liu}]{li2019decomposable}
Zichao Li, Xin Jiang, Lifeng Shang, and Qun Liu. 2019.
\newblock Decomposable neural paraphrase generation.
\newblock In \emph{Proceedings of the 57th Annual Meeting of the Association
  for Computational Linguistics}, pages 3403--3414.

\bibitem[{Liang et~al.(2020)Liang, Yu, Jiang, Er, Wang, Zhao, and
  Zhang}]{liang2020bond}
Chen Liang, Yue Yu, Haoming Jiang, Siawpeng Er, Ruijia Wang, Tuo Zhao, and Chao
  Zhang. 2020.
\newblock Bond: Bert-assisted open-domain named entity recognition with distant
  supervision.
\newblock In \emph{Proceedings of the 26th ACM SIGKDD International Conference
  on Knowledge Discovery \& Data Mining}, pages 1054--1064.

\bibitem[{Lin et~al.(2021)Lin, Zou, and Ding}]{lin2021using}
Jieyu Lin, Jiajie Zou, and Nai Ding. 2021.
\newblock Using adversarial attacks to reveal the statistical bias in machine
  reading comprehension models.
\newblock \emph{arXiv preprint arXiv:2105.11136}.

\bibitem[{Liu et~al.(2019)Liu, Ott, Goyal, Du, Joshi, Chen, Levy, Lewis,
  Zettlemoyer, and Stoyanov}]{liu2019roberta}
Yinhan Liu, Myle Ott, Naman Goyal, Jingfei Du, Mandar Joshi, Danqi Chen, Omer
  Levy, Mike Lewis, Luke Zettlemoyer, and Veselin Stoyanov. 2019.
\newblock \href {http://arxiv.org/abs/1907.11692} {Roberta: {A} robustly
  optimized {BERT} pretraining approach}.
\newblock \emph{CoRR}, abs/1907.11692.

\bibitem[{Loshchilov and Hutter(2018)}]{loshchilov2018decoupled}
Ilya Loshchilov and Frank Hutter. 2018.
\newblock Decoupled weight decay regularization.
\newblock In \emph{International Conference on Learning Representations}.

\bibitem[{Manjunatha et~al.(2019)Manjunatha, Saini, and
  Davis}]{manjunatha2019explicit}
Varun Manjunatha, Nirat Saini, and Larry~S Davis. 2019.
\newblock Explicit bias discovery in visual question answering models.
\newblock In \emph{2019 IEEE/CVF Conference on Computer Vision and Pattern
  Recognition (CVPR)}, pages 9554--9563. IEEE Computer Society.

\bibitem[{Meng et~al.(2021)Meng, Zhang, Huang, Wang, Zhang, Ji, and
  Han}]{meng2021distantly}
Yu~Meng, Yunyi Zhang, Jiaxin Huang, Xuan Wang, Yu~Zhang, Heng Ji, and Jiawei
  Han. 2021.
\newblock Distantly-supervised named entity recognition with noise-robust
  learning and language model augmented self-training.
\newblock In \emph{Proceedings of the 2021 Conference on Empirical Methods in
  Natural Language Processing}, pages 10367--10378.

\bibitem[{Miller(1995)}]{miller1995wordnet}
George~A Miller. 1995.
\newblock Wordnet: a lexical database for english.
\newblock \emph{Communications of the ACM}, 38(11):39--41.

\bibitem[{Palangi et~al.(2016)Palangi, Deng, Shen, Gao, He, Chen, Song, and
  Ward}]{palangi2016deep}
Hamid Palangi, Li~Deng, Yelong Shen, Jianfeng Gao, Xiaodong He, Jianshu Chen,
  Xinying Song, and Rabab Ward. 2016.
\newblock Deep sentence embedding using long short-term memory networks:
  Analysis and application to information retrieval.
\newblock \emph{IEEE/ACM Transactions on Audio, Speech, and Language
  Processing}, 24(4):694--707.

\bibitem[{Pang et~al.(2021)Pang, Lan, and Cheng}]{pang2021match}
Liang Pang, Yanyan Lan, and Xueqi Cheng. 2021.
\newblock Match-ignition: Plugging pagerank into transformer for long-form text
  matching.
\newblock \emph{arXiv preprint arXiv:2101.06423}.

\bibitem[{Papineni et~al.(2002)Papineni, Roukos, Ward, and
  Zhu}]{papineni2002bleu}
Kishore Papineni, Salim Roukos, Todd Ward, and Wei-Jing Zhu. 2002.
\newblock \href {https://doi.org/10.3115/1073083.1073135} {{B}leu: a method for
  automatic evaluation of machine translation}.
\newblock In \emph{Proceedings of the 40th Annual Meeting of the Association
  for Computational Linguistics}, pages 311--318, Philadelphia, Pennsylvania,
  USA. Association for Computational Linguistics.

\bibitem[{Parikh et~al.(2016)Parikh, T{\"a}ckstr{\"o}m, Das, and
  Uszkoreit}]{parikh2016decomposable}
Ankur Parikh, Oscar T{\"a}ckstr{\"o}m, Dipanjan Das, and Jakob Uszkoreit. 2016.
\newblock A decomposable attention model for natural language inference.
\newblock In \emph{Proceedings of the 2016 Conference on Empirical Methods in
  Natural Language Processing}, pages 2249--2255.

\bibitem[{Peng et~al.(2019)Peng, Xing, Zhang, Fu, and
  Huang}]{peng2019distantly}
Minlong Peng, Xiaoyu Xing, Qi~Zhang, Jinlan Fu, and Xuan-Jing Huang. 2019.
\newblock Distantly supervised named entity recognition using
  positive-unlabeled learning.
\newblock In \emph{Proceedings of the 57th Annual Meeting of the Association
  for Computational Linguistics}, pages 2409--2419.

\bibitem[{Pennington et~al.(2014)Pennington, Socher, and
  Manning}]{pennington2014glove}
Jeffrey Pennington, Richard Socher, and Christopher~D Manning. 2014.
\newblock Glove: Global vectors for word representation.
\newblock In \emph{Proceedings of the 2014 conference on empirical methods in
  natural language processing (EMNLP)}, pages 1532--1543.

\bibitem[{R{\"u}ckl{\'e} et~al.(2020)R{\"u}ckl{\'e}, Pfeiffer, and
  Gurevych}]{ruckle2020multicqa}
Andreas R{\"u}ckl{\'e}, Jonas Pfeiffer, and Iryna Gurevych. 2020.
\newblock Multicqa: Zero-shot transfer of self-supervised text matching models
  on a massive scale.
\newblock In \emph{Proceedings of the 2020 Conference on Empirical Methods in
  Natural Language Processing (EMNLP)}, pages 2471--2486.

\bibitem[{Su et~al.(2021)Su, Vandyke, Baker, Wang, and Collier}]{su2021keep}
Yixuan Su, David Vandyke, Simon Baker, Yan Wang, and Nigel Collier. 2021.
\newblock Keep the primary, rewrite the secondary: A two-stage approach for
  paraphrase generation.
\newblock In \emph{Findings of the Association for Computational Linguistics:
  ACL-IJCNLP 2021}, pages 560--569.

\bibitem[{Tai et~al.(2015)Tai, Socher, and Manning}]{tai2015improved}
Kai~Sheng Tai, Richard Socher, and Christopher~D Manning. 2015.
\newblock Improved semantic representations from tree-structured long
  short-term memory networks.
\newblock In \emph{Proceedings of the 53rd Annual Meeting of the Association
  for Computational Linguistics and the 7th International Joint Conference on
  Natural Language Processing (Volume 1: Long Papers)}, pages 1556--1566.

\bibitem[{Tan et~al.(2018)Tan, Wei, Wang, Lv, and Zhou}]{tan2018multiway}
Chuanqi Tan, Furu Wei, Wenhui Wang, Weifeng Lv, and Ming Zhou. 2018.
\newblock Multiway attention networks for modeling sentence pairs.
\newblock In \emph{IJCAI}, pages 4411--4417.

\bibitem[{Tan et~al.(2016)Tan, Dos~Santos, Xiang, and Zhou}]{tan2016improved}
Ming Tan, Cicero Dos~Santos, Bing Xiang, and Bowen Zhou. 2016.
\newblock Improved representation learning for question answer matching.
\newblock In \emph{Proceedings of the 54th Annual Meeting of the Association
  for Computational Linguistics (Volume 1: Long Papers)}, pages 464--473.

\bibitem[{Tay et~al.(2018{\natexlab{a}})Tay, Luu, and Hui}]{tay2018co}
Yi~Tay, Anh~Tuan Luu, and Siu~Cheung Hui. 2018{\natexlab{a}}.
\newblock Co-stack residual affinity networks with multi-level attention
  refinement for matching text sequences.
\newblock In \emph{Proceedings of the 2018 Conference on Empirical Methods in
  Natural Language Processing}, pages 4492--4502.

\bibitem[{Tay et~al.(2018{\natexlab{b}})Tay, Luu, and Hui}]{tay2018compare}
Yi~Tay, Anh~Tuan Luu, and Siu~Cheung Hui. 2018{\natexlab{b}}.
\newblock Compare, compress and propagate: Enhancing neural architectures with
  alignment factorization for natural language inference.
\newblock In \emph{Proceedings of the 2018 Conference on Empirical Methods in
  Natural Language Processing}, pages 1565--1575.

\bibitem[{Vaswani et~al.(2017)Vaswani, Shazeer, Parmar, Uszkoreit, Jones,
  Gomez, Kaiser, and Polosukhin}]{vaswani2017attention}
Ashish Vaswani, Noam Shazeer, Niki Parmar, Jakob Uszkoreit, Llion Jones,
  Aidan~N Gomez, {\L}ukasz Kaiser, and Illia Polosukhin. 2017.
\newblock Attention is all you need.
\newblock In \emph{Advances in neural information processing systems}, pages
  5998--6008.

\bibitem[{Wang et~al.(2019{\natexlab{a}})Wang, Pruksachatkun, Nangia, Singh,
  Michael, Hill, Levy, and Bowman}]{wang2019superglue}
Alex Wang, Yada Pruksachatkun, Nikita Nangia, Amanpreet Singh, Julian Michael,
  Felix Hill, Omer Levy, and Samuel~R Bowman. 2019{\natexlab{a}}.
\newblock Superglue: a stickier benchmark for general-purpose language
  understanding systems.
\newblock In \emph{Proceedings of the 33rd International Conference on Neural
  Information Processing Systems}, pages 3266--3280.

\bibitem[{Wang et~al.(2019{\natexlab{b}})Wang, Singh, Michael, Hill, Levy, and
  Bowman}]{wang2019glue}
Alex Wang, Amanpreet Singh, Julian Michael, Felix Hill, Omer Levy, and Samuel~R
  Bowman. 2019{\natexlab{b}}.
\newblock Glue: A multi-task benchmark and analysis platform for natural
  language understanding.
\newblock In \emph{7th International Conference on Learning Representations,
  ICLR 2019}.

\bibitem[{Wang et~al.(2019{\natexlab{c}})Wang, Jiang, Xu, and
  Zhang}]{wang2019knowledge}
Peilu Wang, Hao Jiang, Jingfang Xu, and Qi~Zhang. 2019{\natexlab{c}}.
\newblock Knowledge graph construction and applications for web search and
  beyond.
\newblock \emph{Data Intelligence}, 1(4):333--349.

\bibitem[{Wang et~al.(2021)Wang, Liu, Gui, Zhang, Zou, Zhou, Ye, Zhang, Zheng,
  Pang et~al.}]{wang2021textflint}
Xiao Wang, Qin Liu, Tao Gui, Qi~Zhang, Yicheng Zou, Xin Zhou, Jiacheng Ye,
  Yongxin Zhang, Rui Zheng, Zexiong Pang, et~al. 2021.
\newblock Textflint: Unified multilingual robustness evaluation toolkit for
  natural language processing.
\newblock In \emph{Proceedings of the 59th Annual Meeting of the Association
  for Computational Linguistics and the 11th International Joint Conference on
  Natural Language Processing: System Demonstrations}, pages 347--355.

\bibitem[{Wang et~al.(2017)Wang, Hamza, and Florian}]{wang2017bilateral}
Zhiguo Wang, Wael Hamza, and Radu Florian. 2017.
\newblock Bilateral multi-perspective matching for natural language sentences.
\newblock In \emph{Proceedings of the 26th International Joint Conference on
  Artificial Intelligence}, pages 4144--4150.

\bibitem[{Wang et~al.(2016)Wang, Mi, and Ittycheriah}]{wang2016sentence}
Zhiguo Wang, Haitao Mi, and Abraham Ittycheriah. 2016.
\newblock Sentence similarity learning by lexical decomposition and
  composition.
\newblock In \emph{Proceedings of COLING 2016, the 26th International
  Conference on Computational Linguistics: Technical Papers}, pages 1340--1349.

\bibitem[{Wu et~al.(2020)Wu, Yang, Zhang, Hong, Fu, and Wang}]{wu2020joint}
Le~Wu, Yonghui Yang, Kun Zhang, Richang Hong, Yanjie Fu, and Meng Wang. 2020.
\newblock Joint item recommendation and attribute inference: An adaptive graph
  convolutional network approach.
\newblock In \emph{Proceedings of the 43rd International ACM SIGIR Conference
  on Research and Development in Information Retrieval}, pages 679--688.

\bibitem[{Yang et~al.(2019)Yang, Zhang, Gao, Ji, and Chen}]{yang2019simple}
Runqi Yang, Jianhai Zhang, Xing Gao, Feng Ji, and Haiqing Chen. 2019.
\newblock Simple and effective text matching with richer alignment features.
\newblock In \emph{Proceedings of the 57th Annual Meeting of the Association
  for Computational Linguistics}, pages 4699--4709.

\bibitem[{Zhang et~al.(2017)Zhang, Chen, Liu, Liu, and Lv}]{zhang2017context}
Kun Zhang, Enhong Chen, Qi~Liu, Chuanren Liu, and Guangyi Lv. 2017.
\newblock A context-enriched neural network method for recognizing lexical
  entailment.
\newblock In \emph{Proceedings of the AAAI Conference on Artificial
  Intelligence}, volume~31.

\bibitem[{Zhang et~al.(2019)Zhang, Lv, Wang, Wu, Chen, Wu, and
  Xie}]{zhang2019drr}
Kun Zhang, Guangyi Lv, Linyuan Wang, Le~Wu, Enhong Chen, Fangzhao Wu, and Xing
  Xie. 2019.
\newblock Drr-net: Dynamic re-read network for sentence semantic matching.
\newblock In \emph{Proceedings of the AAAI Conference on Artificial
  Intelligence}, volume~33, pages 7442--7449.

\bibitem[{Zhang et~al.(2021)Zhang, Wu, Lv, Wang, Chen, and
  Ruan}]{zhang2021making}
Kun Zhang, Le~Wu, Guangyi Lv, Meng Wang, Enhong Chen, and Shulan Ruan. 2021.
\newblock Making the relation matters: Relation of relation learning network
  for sentence semantic matching.
\newblock In \emph{Proceedings of the AAAI Conference on Artificial
  Intelligence}, volume~35, pages 14411--14419.

\end{thebibliography}
\bibliographystyle{acl_natbib}

\end{document}